\documentclass[11pt]{article}

\usepackage{setspace} 
\usepackage{amsmath}
\usepackage{graphicx,psfrag,epsf}
\usepackage{enumerate}
\usepackage{natbib}
\usepackage{url}
\usepackage{ amssymb, amsthm, bm}
\usepackage{bmpsize}
\usepackage[figuresright]{rotating}
\usepackage{psfrag, epsf}
\usepackage{enumerate}
\usepackage{lineno}
\modulolinenumbers[5]
\usepackage{booktabs}
\usepackage[utf8]{inputenc}
\usepackage{graphicx}
\usepackage{natbib}
\usepackage[breaklinks]{hyperref}
\usepackage{breakurl}
\usepackage{flafter}
\usepackage{float}
\usepackage{adjustbox}
\usepackage{tgbonum}
\newcommand{\blind}{0}

\usepackage{multicol}
\usepackage{adjustbox}
\usepackage{cclicenses}
\usepackage{makecell}
\usepackage{lscape,array}
\newcolumntype{C}[1]{>{\centering\arraybackslash}p{#1}} 
\usepackage[thin, , thinc]{esdiff}
\usepackage{subcaption}
\usepackage{placeins}
\usepackage{xcolor}
\usepackage{istgame}
\usepackage{wrapfig}
\usepackage{bmpsize}
\usepackage{flafter}
\usepackage{color, hyperref, xcolor}
\hypersetup{colorlinks=true, urlcolor=blue, citecolor=purple}
\usepackage{cleveref}
\numberwithin{equation}{section}

\newtheoremstyle{general}
{3mm} % Space above
{3mm} % Space below
{} % Body font
{} % Indent amount
{\bfseries} % Theorem head font
{.} % Punctuation after theorem head
{.5em} % Space after theorem head
{} % Theorem head spec (can be left empty,  meaning `normal')

\theoremstyle{general}

\begin{document}

\doublespacing

%%%%%%%%%%%%%%%%%%%%%%%%%%%%%%%%%%%%%%%%%%%%%%%%%%%%%%%%%%%%%%%%%%%%%%%%%%%%%%

\if0\blind
{
\title{On Language Clustering: A Non-parametric Statistical Approach}
  \author{ Anagh Chattopadhyay$^{1}$\thanks{Corresponding author, email: anagh72@gmail.com},Soumya Sankar Ghosh$^{2}$ \& Samir Karmakar$^{3}$  \vspace{.1cm}\\
 $^{1}$  Indian Statistical Institute, 203 B.T. Road, Kolkata, India.\vspace{.1cm}\\
 $^{2}$ VIT Bhopal University, Bhopal, India.\vspace{.1cm}\\
 $^{3}$ Jadavpur University, Kolkata, India.
}
  \maketitle
} \fi

\if1\blind
{
  \bigskip
  \bigskip
  \begin{center}
    {\LARGE\bf Analysis and Estimation of Consumer Expenditure assuming Uniform Prices across FSUs:  A step towards Cost Effectiveness}
\end{center}
  \medskip
} \fi

\bigskip
\begin{abstract}
Any approach aimed at pasteurizing and quantifying a particular phenomenon must include the use of robust statistical methodologies for data analysis. With this in mind, the purpose of this study is to present statistical approaches that may be employed in nonparametric nonhomogeneous data frameworks, as well as to examine their application in the field of natural language processing and language clustering. Furthermore, this paper discusses the many uses of nonparametric approaches in linguistic data mining and processing. The data depth idea allows for the centre-outward ordering of points in any dimension, resulting in a new nonparametric multivariate statistical analysis that does not require any distributional assumptions. The concept of hierarchy is used in historical language categorisation and structuring, and it aims to organise and cluster languages into subfamilies using the same premise.
In this regard, the current study presents a novel approach to language family structuring based on non-parametric approaches produced from a typological structure of words in various languages, which is then converted into a Cartesian framework using MDS. This statistical-depth-based architecture allows for the use of data-depth-based methodologies for robust outlier detection, which is extremely useful in understanding the categorization of diverse borderline languages and allows for the re-evaluation of existing classification systems. Other depth-based approaches are also applied to processes such as unsupervised and supervised clustering. This paper therefore provides an overview of procedures that can be applied to nonhomogeneous language classification systems in a nonparametric framework.
\end{abstract}

\noindent%
\vfill

\section{Introduction}

The goal of data clustering is to divide a set of \textit{n} items into groups, which can be represented as points in a \textit{d} dimensional space or as a \textit{n*n} similarity matrix. Due to the lack of a common definition of a cluster and its task or data-dependent nature, a great number of clustering algorithms have been published, each with various assumptions regarding cluster formation. The proposed methodologies can be divided into two categories: parametric and non-parametric approaches. Non-parametric methods , as the name sugegsts does not assume any distrubutional assumptions about the data, whereas parametric approaches does. \cite{siegel1957nonparametric} and \cite{savage1957nonparametric} introduced this nonparametric model in many forms for applications, while \cite{wasserman2006all} provides a complete study of nonparametric approaches in the statistical paradigm. This mentioned model, particularly, does not require any data or data point distributional assumptions, allowing it to give considerably more versatile and robust inference and classification approaches that may be applied to a wide range of complex or limited data structures. When the data given isn't substantial enough to deduce distributional qualities from or isn't suitable for applying distributional assumptions, non-parametric statistics is of special interest.
The current research project focuses on using this non-parametric model in natural language clustering to better understand how languages are spread around the world while maintaining linguistic convergence. To achieve this goal, the paper will be divided into four sections, in which Section $2$ will focus primarily on the concept of data depth and its relation to linguistics. In Section $3$ the discussion will be further augmented with the data and methodological part. Finally, in Section $4$, the paper will analyze the framework that is crucial in providing a systemic account of language clustering.

%On the other hand, language clustering necessitates a complex structural network. It needs to create a classification system for natural languages based on their overall similarity. The main problem in this classification stems from the internal structure of the information and the (dis)similarities of these internal structures. Since obtaining the distributional properties of languages or language families and their distributions is difficult, turning to non-parametric statistics and their application is a natural solution to this problem.

\section{Data Depth and Linguistic Clustering }
The nonparametric approach to multivariate data analysis includes a concept called data depth. It gives one method to organise multivariate data. This type of order is known as central-outward ordering. A depth function is essentially any function that offers a ``reasonable" central-outward ordering of points in multidimensional space. How "close" a particular point is some measure of center of a distribution is the principle idea behind data depth. Many other depth-based approaches have been described in the literature, including multivariate paradigms (\cite{liu1999multivariate}, \cite{vardi2000multivariate}, \cite{he1997convergence}) and univariate paradigms (\cite{dyckerhoff1996zonoid}, \cite{aloupis2006geometric}, etc.). \cite{lange2014fast} makes a remark on nonparametric classification based on data depth and suggests a fast classification approach based on nonparametric statistics. According to \cite{zuo2000general}, statistical depth is a function that has the following:
	\begin{itemize}
\item affine transformation invariance
\item maximality at the center of symmetry of the distribution for the class of symmetric distributions
\item monotonicity relative to the point with the highest depth
\item vanishing at infinity
\end{itemize}
This will result in a function that recognizes ``typical" and ``outlier" observations, as well as a quantile generalization for multivariate data. Because of its non-parametric nature, data depth allows a great deal of freedom for data analysis and categorization and is therefore quite general. It is a notion that provides an alternate idea or methodology for assessing the center/centrality in a data frame, as well as a measure of centrality.

\begin{figure}[htp]
    \centering
    \includegraphics[width=\textwidth]{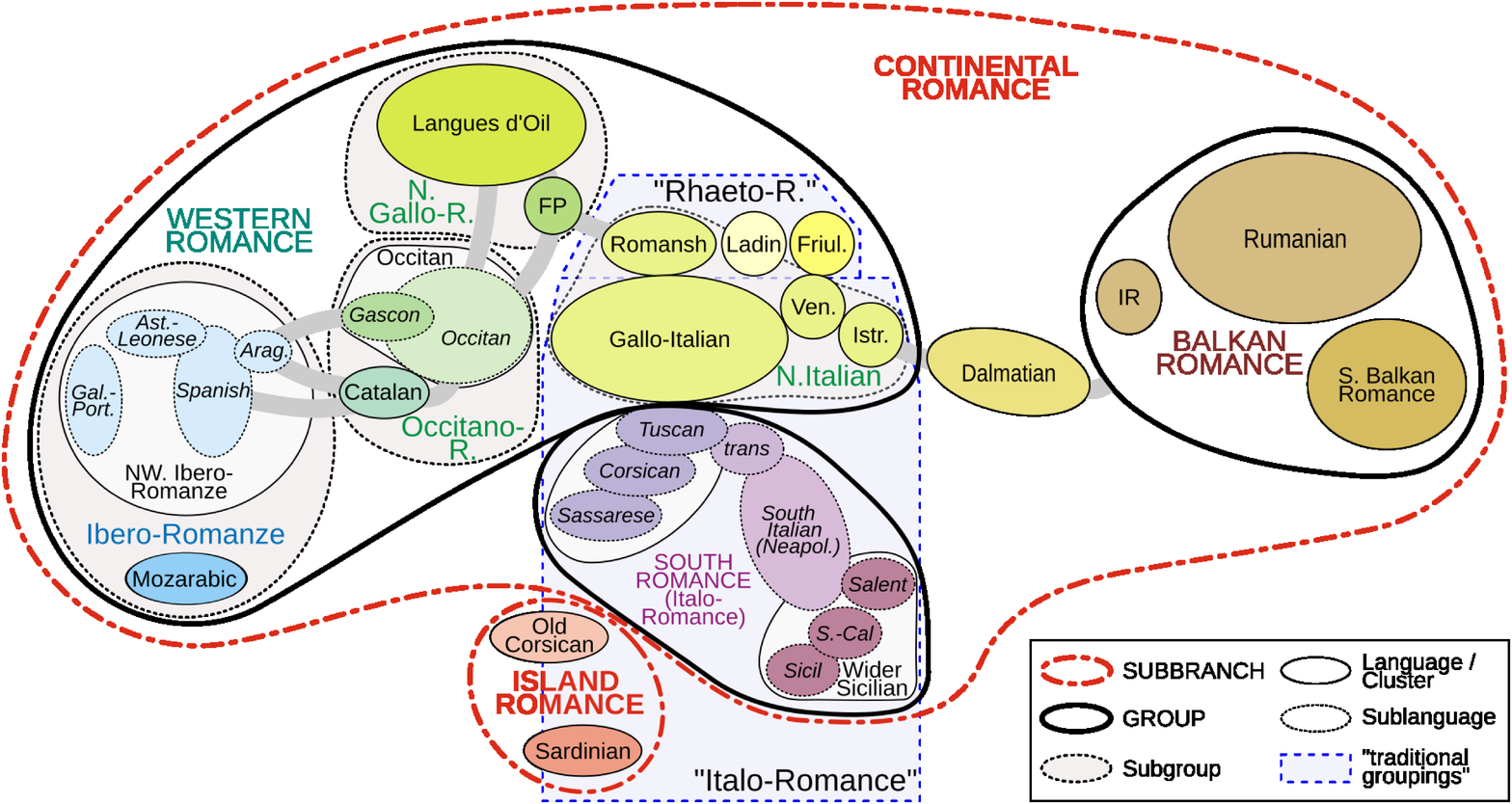}
   \caption{Chart of Romance languages based on structural and comparative criteria, not on socio-functional ones. FP: Franco-Provençal, IR: Istro-Romanian}
    \label{fig2}
    \end{figure}

It is this data depth property of being able to distinguish between ``typical" and ``outlier" observations that helps us achieve our goal of outlier detection in language families, as well as the general nonparametric properties and tools that can be used to study the structure of the various languages in a ``language space," which is especially useful because we don't know any distributional properties of the languages in that space. Nonparametric approaches, as their name implies, do not require data to be parameterized, and are thus particularly effective in situations like this, where no distributional structure is present.
We have considered Romance languages for the sake of this study, which are a subgroup of the italic branch of the Indo-European language family. The main languages of this language group are French, Italian, Spanish, Portuguese, and Romanian. For a better understanding, consider Fig. \ref{fig2}

%\subsection{Data Collection}
   
%The Romance family is quite versatile,and the language family hierarchy is given below:-

%Romance languages, group of related languages all derived from Vulgar Latin within historical times and forming a subgroup of the Italic branch of the Indo-European language family. The major languages of the family include French, Italian, Spanish, Portuguese, and Romanian, all national languages. The romance language group has been focused upon primarily in this paper.

\section{Data Collection and Methodology}

The overall methodology in this paper can be understood in the following step-by-step manner:-
\begin{enumerate}
    \item We begin by creating sets of the most frequently used terms in the languages we are interested in (in our case, Romance languages) that correlate to a set of fixed and common meanings.
    \item The distance matrix relating to the distance between any pair of languages is then calculated using the Levenshtein distance metric. This stage lays the groundwork for creating a language space with the right amount of dimensions. The use of various non-parametric approaches is based on this language space.
    \item Dimension scaling, which is done using MDS (multidimensional scaling), is the next stage in creating the language space, and then we embed the points belonging to various languages in an abstract Cartesian system (with appropriate scaling measures). 
    \item Finally, we use appropriate non-parametric measures to analyze the numerous outlier-based features of this particular structure. 
\end{enumerate}

The following subsections deal with each of these steps in the aforementioned order.

\subsection{Data collection}

 The primary step is the collection of data,which has been done primarily from the \textit{Serva} corpus \cite{serva} and from the database of \cite{swadesh1955towards} and \cite{wichmann2011phonological}.The \textit{Serva} corpus basically contains the most common word corresponding to a given language and a common meaning particularly for the romance languages. This exhaustive list contains 60 languages and about 110 words. A small part of the data set has been shown in table \ref{fig3}. We have primarily focused on Romance languages. Although using a single word meaning and understanding the typological differences between the corresponding words in various languages may be useful for understanding localised evolutionary properties corresponding to that word meaning, it causes a lot of spurious errors when used to understand the overall differences between different languages as entire entities. However, this randomness, which may be evened out by utilising a large number of words, can be reliably extracted to a substantial extent, resulting in a robust grasp of the language space. 

\begin{table}[h!]
\centering
\begin{tabular}{|c|c |c |c |c |c|} 
  \hline
Word & Classical.Latin & Meg.Romanian & Ist.Romanian & Aromanian & Romanian \\ 
  \hline
 all & omnis & tot & tot & tut & tot \\ 
  ashes & cinis & tsanusa & ceruse & cinuse & cenusa \\ 
 bark & cortex & coaja & cora & coaje & scoarta \\ 
belly & venter & foali & tarbuh & pintic & burta \\ 
 big & grandis & mari & mare & mare & mare \\ 
 bird & avis & pul & pul & puliu & pasare \\ 
 bite & mordere & mutscu & mucca & miscu & musca \\ 
   \hline
\end{tabular}
\caption{A peek of the data set}
\label{fig3}
\end{table}

\subsection{Distance matrix}
A string metric for quantifying the difference between two sequences is the Levenshtein distance, which is a sort of edit distance. Following this, the levenshtein distance between two words is defined as the minimum number of single edits that  possess the ability to change one word into the other(\cite{levenshtein1966binary}). It has been widely used in the linguistic literature for phrasal and typological differentiation in a variety of contexts, such as in the works of \cite{nerbonne1999edit}, \cite{nerbonne1997measuring}, \cite{nerbonne1996phonetic}, and others. It has been used in these works to quantify dialect differences and to measure phonetic differences in languages.

The Levenshtein distance between two strings p and q  (of length $|p|$ and $|q|$ respectively) which is written as lev$(p, q)$ and is defined in \ref{eq-1}

\small
\begin{equation}
\label{eq-1}
\text{levenshtein.dist(p,q)}= \begin{cases}
|p| &\text{if $|q|=0$}\\
|q| &\text{if $|p|=0$}\\
\text{levenshtein.dist(tail(p),tail(q))} &\text{if head(p)=head(q)}\\
1+\text{min} \begin{cases}
\text{levenshtein.dist(tail(p),q)}\\
\text{levenshtein.dist(p,tail(q))}\\
\text{levenshtein.dist(p,q)}
\end{cases}
 &\text{otherwise}

\end{cases}
\end{equation}

In \ref{eq-1} \textit{head} the first letter of the word is denoted, while in \textit{tail} the rest of the word is indicated after the head is removed. This distance forms the pivotal entity that helps form the distance matrix corresponding to the languages.

\normalsize
    We compute the Levenshtein distance between all language pairings for a given word meaning and provide a matrix that matches the word meaning. The matrix is a symmetric matrix with all of the diagonal elements equal to zero, as one might expect. We obtain a different Levenshtein matrix for the meaning of each word. Language-based differentiation can be done at two levels: a localized level, which looks at each Levenshtein distance matrix corresponding to a specific word meaning, and a globalized level, which is an algebraic function of all the Levenshtein distances obtained across all word meanings in all samples. Due to the general rationale below, we will concentrate on the aforementioned globalized structure.

 It has already mentioned that using a single word meaning to create the distance matrix can lead to incorrect results, especially because the similarity structure of words corresponding to a single meaning can show disproportionate similarity and dissimilarity among languages far and near from each other, especially due to the dominance of chance causes of similarity (or dissimilarity).

As a result, we average the distance matrices obtained from the multiple word meanings used, yielding a sensible and robust distance matrix that captures the similarity structure between the languages.

The final distance matrix obtained is shown in table \ref{fig34}

\begin{table}[h!]
\centering
\begin{tabular}{||c||c|c|c|c|} 
  \hline
 & Late.Clas.Latin & Meg.Romanian & Is.Romanian & Aromanian \\ 
  \hline
Late.Classical.Latin & 0.00 & 3.69 & 3.73 & 3.67 \\ 
  Megleno.Romanian & 3.69 & 0.00 & 2.44 & 1.76 \\ 
  Istro.Romanian & 3.73 & 2.44 & 0.00 & 2.39  \\ 
  Aromanian & 3.67 & 1.76 & 2.39 & 0.00  \\ 
  
   \hline
\end{tabular}
\caption{A principal submatrix corresponding to the obtained distance matrix}
\label{fig34}
\end{table}

\subsection{On Law of large numbers and sampling procedures}

This property corresponding to the lack of interpretability of distance matrices when considering just a few common words, and the subsequent intepretability and validity of the same in the cases of consideration over a large number of common words can be induced or derived with respect to an analogous form of the law of large numbers, because in an appropriate language space corresponding to the various languages, if $\theta_i$ refers to the position of the $i^{th}$ language in our Cartesian plane, with  \[\delta_{ij}=\theta_i-\theta_j
 \]
 
 referring to the population parameter corresponding to the average Levenstein distance between the most common word corresponding to the two languages with a common meaning, averaged over all common meanings possible, and letting $\hat{\delta}_{ij,n}$,$\hat{\theta}_{i,n}$,$\hat{\theta}_{j,n}$ to be the corresponding sample counterparts obtained from the our collected data with say $n$ elements, then by the law of large number, 
 
 \[
 \hat{\delta}_{ij,n}\to \delta_{ij} \hspace{4mm} \text{as} \hspace{4mm} n \to \infty
 \].
 
 A reader who is interested in learning more about the law of large numbers might look up \cite{feller1957introduction}. Since any two well-established, well-spoken, and recognised languages have a large number of common words in most senses, the law of large numbers motivates us to select a large number of common words between the languages of interest in order to derive robust and interpretable results with reasonable levels of error.
    
\subsection{Interpreting the distance matrix}
\begin{figure}[htp]
    \centering
    \includegraphics[width=1.4\textwidth, angle =270]{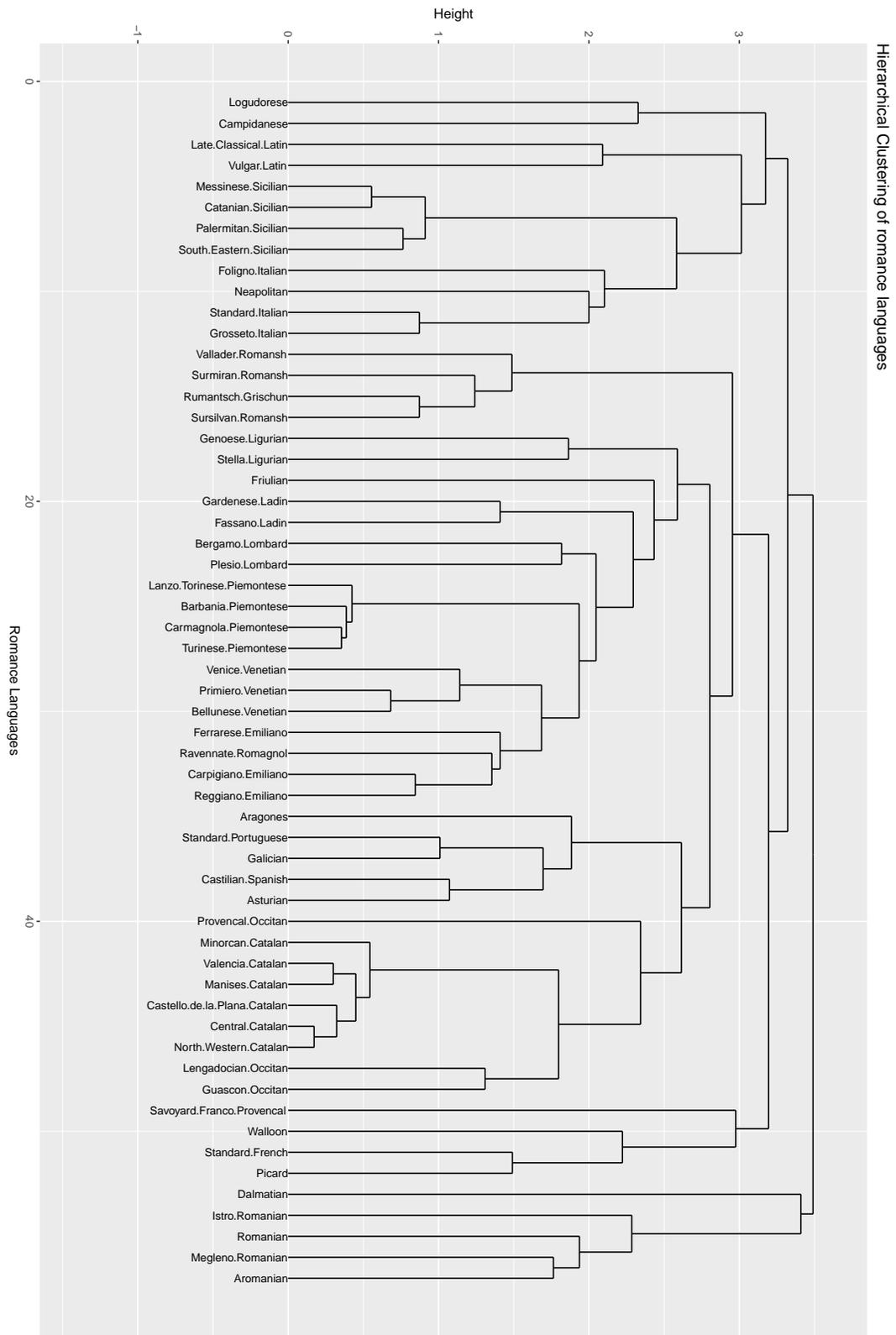}
    \caption{{Hierarchical clustering(we note the lack of any measure of centricity(which can be useful in an Historical Linguistic point of view})(Average method)}
    \label{fig5}
    \end{figure}

\begin{figure}[htp]
    \centering
    \includegraphics[width=1.4\textwidth, angle =270]{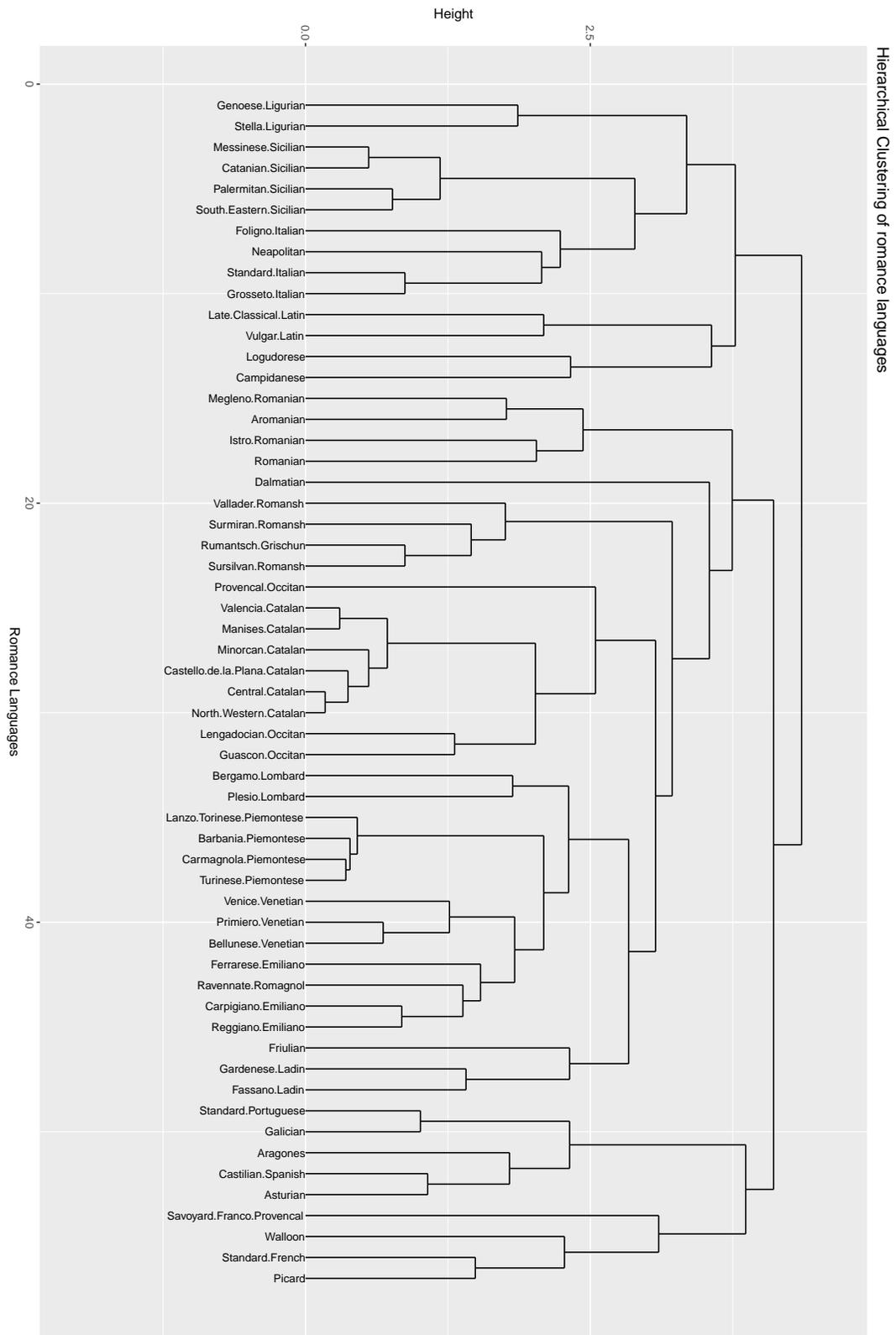}
    \caption{{Hierarchical clustering(we note the lack of any measure of centricity(which can be useful in an Historical Linguistic point of view})(Complete method)}
    \label{fig51}
    \end{figure}

A comprehensive hierarchical clustering is used to check the authenticity and meaning of the produced distance matrix. The results of which are shown in fig \ref{fig5} (corresponding to average linkage clustering) and  fig \ref{fig51} (corresponding to complete linkage clustering).

In complete-link (or complete linkage) hierarchical clustering, we merge the two clusters with the least merger diameter (or: the two clusters with the smallest maximum pairwise distance).

In single-link (or single linkage) hierarchical clustering, we merge the two clusters with the least distance between their two closest members in each step (or: the two clusters with the smallest minimum pairwise distance).

We note that the obtained hierarchical clustering based dendrogram encapsulates the actual structure of the romance family of languages to a huge degree(and we also observe that various dialects of the same language are close), and that helps us verify the validity of the distance matrix , as it becomes important and crucial in the steps mentioned in the succeeding sections.
A comparison of  fig \ref{fig2} with fig \ref{fig5} and fig \ref{fig51} at this point will reveal that the results are very similar, despite the fact that fig \ref{fig2} was obtained using historical procedures. This leads to the current study's conclusion that the law of large numbers is useful in the context of linguistic clustering. It is worth noting that the proposed methodology is capable of recovering a significant amount of information on not only their typological but also historical similarities. As a result, the similarity between observed and obtained hierarchies informs us that this structure captures a significant amount of information, implying that any further use of non-parametric analytical approaches is justified. 

Now, given the language distance matrix, we must attempt to not only visualise the data set, specifically the language placement, but also to embed them on a Cartesian plane. As a result, we use multidimensional scaling with the proper dimensions (obtained by scaling). R provides the essential framework for both classical and non-parametric MDS to be used.

\begin{figure}[htp]
    \centering
    \includegraphics[width=\textwidth]{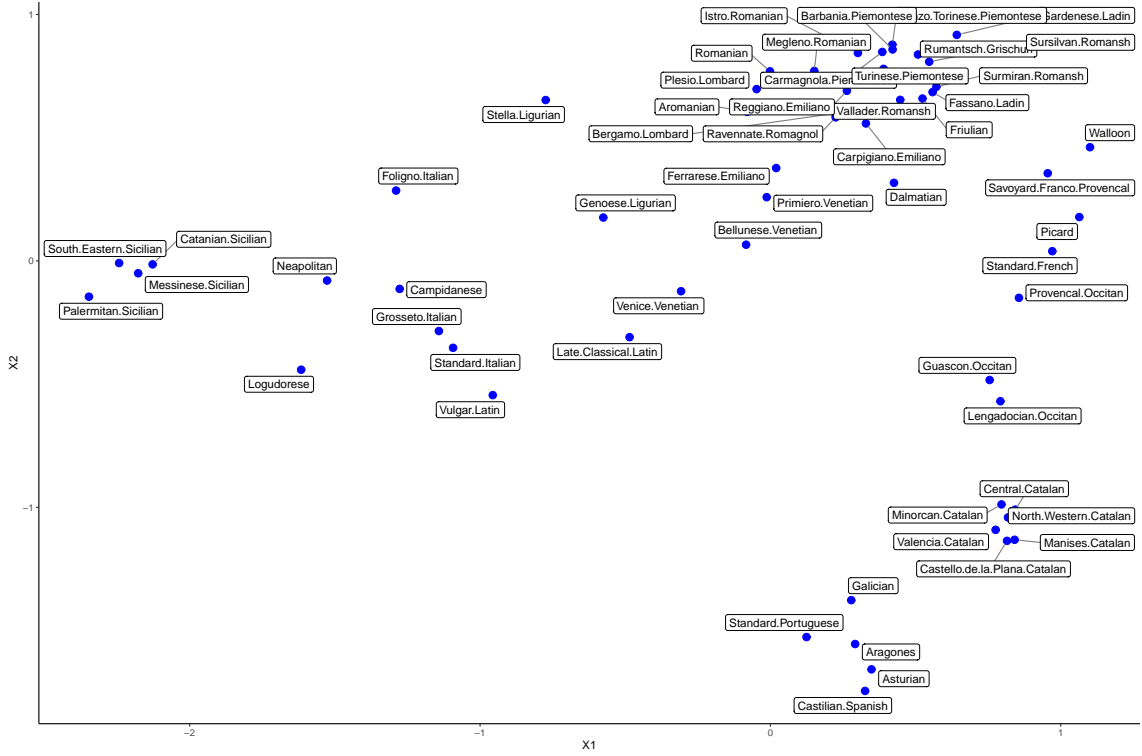}
    \caption{embedding in a Cartesian plane by using MDS}
    \label{fig6}
    \end{figure}

The 2 dimensional MDS for the set of languages is shown in fig \ref{fig6}

\section{Application and Discussion}

We were able to develop and describe a structure consisting of embedded points in a two-dimensional Cartesian plane corresponding to Romance languages in the previous section, which may be used to apply various non-parametric statistical approaches. As the embedding preserves the distance between the various languages in the Romance family, the next subsections show several uses of non-parametric techniques that are important.

\subsection{Outlier Detection}
The application of data-depth based approaches to language group distinction is one of the most essential aspects of their use.
This essentially aids us in developing a statistically sound methodology for determining whether a language belongs to a specific family based on the degree of typological distinctions. We use the \texttt{Spatial.Depth} functionality at R for outlier detection,especially for languages which cannot be easily put into a given language family. We used the above functionality at R,and we added the languages \textit{Hindi} and \textit{Sanskrit} to the list of romance languages. We note that functionality detected the two languages as outliers at 0.05 level.
    
    % \begin{figure}[htp]
    % \centering
    % \includegraphics[width=\textwidth]{snip 7.PNG}
    % \caption{Result obtained from spatial.outlier functionality in R}
    % \label{fig7}
    % \end{figure}
    
\begin{figure}[!htp]
    \centering
    \includegraphics[width=0.6\textwidth]{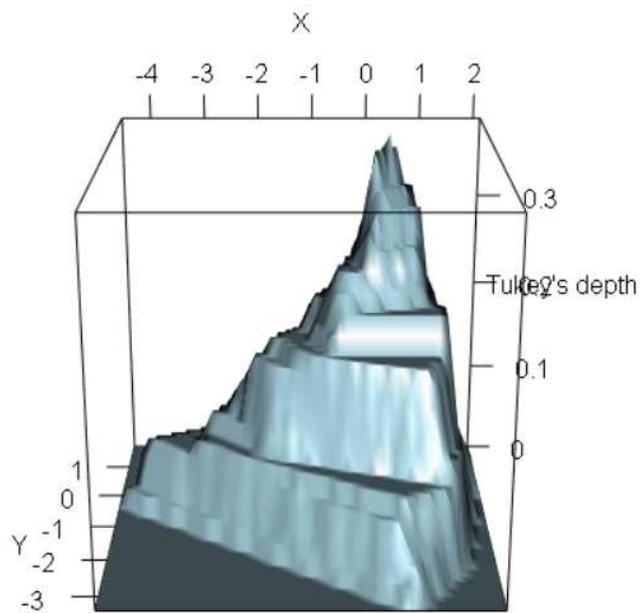}
    \caption{A 3-D perspective of the depth structure of the Languages}
    \label{fig9}
    \end{figure}

\begin{figure}[!htp]
    \centering
    \includegraphics[width=0.6\textwidth]{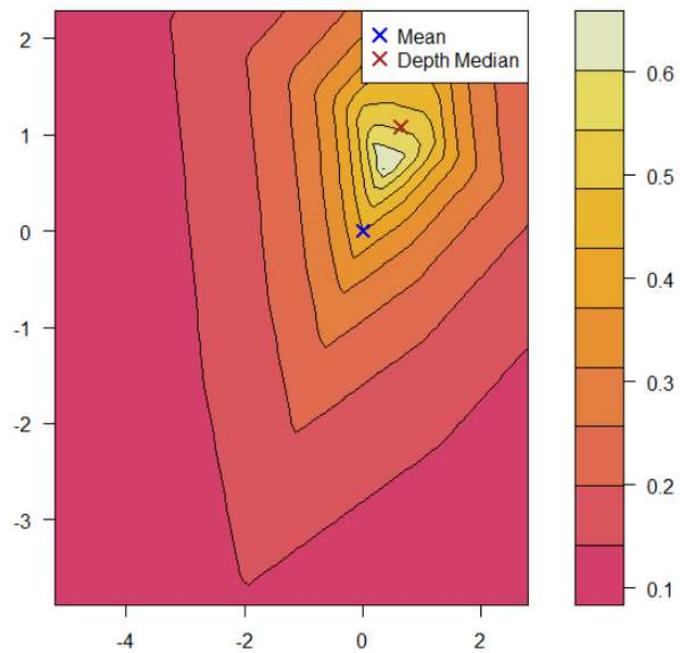}
    \caption{A 2-D heat-map of the depth structure of the Languages}
    \label{fig10}
    \end{figure}

The depth structure of the space of languages is shown in fig \ref{fig9} and \cref{fig10}
  
\subsection{Unsupervised L1-depth based clustering}
Depth based clustering has been studied and is usually implemented for its robustness and intepretability \cite{jeong2016data}
    
We employ the L-1 depth in this case for the clustering process,and use the functionality \texttt{TDDclust} in R which is the trimmed version of the clustering algorithm based on the L1 depth proposed by \cite{jornsten2004clustering}.The paper segments all the observations in clusters, and assigns to each point z in the data space, the L1 depth value regarding its cluster. A trimmed procedure is incorporated to remove the more extreme individuals of each cluster (those one with the lowest depth values).
This methodology allows us to decide the number of clusters we deem suitable for the given language structure,and subsequently allow us the flexibility to cluster,w.r.t the various romance language sub-group structures in the literature.
The result for the case of three clusters centred at \textit{Campidanese}, \textit{Vallader Romansh}, and \textit{Valencia Catalan}.
Application of k-NN non-parametric Regression and PAM was also done, which yielded similar results. Each of these methodologies yield insights towards the inherent groups within the Romance family itself.
	
% 	\begin{figure}[htp]
%     \centering
%     \includegraphics[width=\textwidth]{snip 8.PNG}
%     \caption{Cluster medians obtained from the aforementioned process(taking number of clusters=3)}
%     \label{fig8}
%     \end{figure}
    
\subsection{Supervised Classifications and other possible generalisations}
Depth based Supervised classifications can also be done by training with appropriate languages based training data (which comprised about 80\% of the total data and the remaining 20\% is used for the testing) corresponding to each of the clusters in the proposed classification structure of the languages.
	
We use the package \texttt{ddalpha},which provide framework for the utilisation of the inputted training data,and classifies the test-data w.r.t various types of depth. 82.3\% classification rate was obtained, when we classified among Eastern and Western Romance languages.

  \section{Conclusion} 
 This research demonstrates non-parametric approaches that can aid in the understanding of language families (in our case the romance languages). The paper develops a statistically sound strategy for detecting outliers in the Indo-European family of languages, as well as distinguishing Hindi and Sanskrit from Romance languages. This work also looks at supervised and unsupervised clustering of linguistic sub-families using a non-parametric approach, with promising findings. The success of these non-parametric approaches demonstrates the utility of a Levenshtein distance-based structure, which is not only computationally efficient but also delivers a wealth of information on the behaviour and evolution of different languages with only a modestly high sample size.

\vspace{5mm}
\noindent {\bf Funding source:} 
This research did not receive any specific grant from funding agencies in the public,  commercial,  or not-for-profit sectors. \\      

\noindent {\bf Declarations of interest:} 
The authors declare no conflict of interest.

\bibliographystyle{chicago}
\bibliography{Bibliography-MM-MC}

\begin{thebibliography}{}

\bibitem[\protect\citeauthoryear{Aloupis}{Aloupis}{2006}]{aloupis2006geometric}
Aloupis, G. (2006).
\newblock Geometric measures of data depth.
\newblock {\em DIMACS series in discrete mathematics and theoretical computer
  science\/}~{\em 72}, 147.

\bibitem[\protect\citeauthoryear{Dyckerhoff, Mosler, and Koshevoy}{Dyckerhoff
  et~al.}{1996}]{dyckerhoff1996zonoid}
Dyckerhoff, R., K.~Mosler, and G.~Koshevoy (1996).
\newblock Zonoid data depth: Theory and computation.
\newblock In {\em COMPSTAT}, pp.\  235--240. Springer.

\bibitem[\protect\citeauthoryear{Feller}{Feller}{}]{feller1957introduction}
Feller, W.
\newblock An introduction to probability theory and its applications.
\newblock {\em 1957\/}.

\bibitem[\protect\citeauthoryear{He and Wang}{He and
  Wang}{1997}]{he1997convergence}
He, X. and G.~Wang (1997).
\newblock Convergence of depth contours for multivariate datasets.
\newblock {\em The Annals of Statistics\/}, 495--504.

\bibitem[\protect\citeauthoryear{Jeong, Cai, Sullivan, and Wang}{Jeong
  et~al.}{2016}]{jeong2016data}
Jeong, M.-H., Y.~Cai, C.~J. Sullivan, and S.~Wang (2016).
\newblock Data depth based clustering analysis.
\newblock In {\em Proceedings of the 24th ACM SIGSPATIAL International
  Conference on Advances in Geographic Information Systems}, pp.\  1--10.

\bibitem[\protect\citeauthoryear{J{\"o}rnsten}{J{\"o}rnsten}{2004}]{jornsten2004clustering}
J{\"o}rnsten, R. (2004).
\newblock Clustering and classification based on the l1 data depth.
\newblock {\em Journal of Multivariate Analysis\/}~{\em 90\/}(1), 67--89.

\bibitem[\protect\citeauthoryear{Lange, Mosler, and Mozharovskyi}{Lange
  et~al.}{2014}]{lange2014fast}
Lange, T., K.~Mosler, and P.~Mozharovskyi (2014).
\newblock Fast nonparametric classification based on data depth.
\newblock {\em Statistical Papers\/}~{\em 55\/}(1), 49--69.

\bibitem[\protect\citeauthoryear{Levenshtein et~al.}{Levenshtein
  et~al.}{1966}]{levenshtein1966binary}
Levenshtein, V.~I. et~al. (1966).
\newblock Binary codes capable of correcting deletions, insertions, and
  reversals.
\newblock In {\em Soviet physics doklady}, Volume~10, pp.\  707--710. Soviet
  Union.

\bibitem[\protect\citeauthoryear{Liu, Parelius, and Singh}{Liu
  et~al.}{1999}]{liu1999multivariate}
Liu, R.~Y., J.~M. Parelius, and K.~Singh (1999).
\newblock Multivariate analysis by data depth: descriptive statistics, graphics
  and inference,(with discussion and a rejoinder by liu and singh).
\newblock {\em The annals of statistics\/}~{\em 27\/}(3), 783--858.

\bibitem[\protect\citeauthoryear{Nerbonne and Heeringa}{Nerbonne and
  Heeringa}{1997}]{nerbonne1997measuring}
Nerbonne, J. and W.~Heeringa (1997).
\newblock Measuring dialect distance phonetically.
\newblock In {\em Computational phonology: third meeting of the acl special
  interest group in computational phonology}.

\bibitem[\protect\citeauthoryear{Nerbonne, Heeringa, and Kleiweg}{Nerbonne
  et~al.}{1999}]{nerbonne1999edit}
Nerbonne, J., W.~Heeringa, and P.~Kleiweg (1999).
\newblock Edit distance and dialect proximity.
\newblock {\em Time Warps, String Edits and Macromolecules: The theory and
  practice of sequence comparison\/}~{\em 15}.

\bibitem[\protect\citeauthoryear{Nerbonne, Heeringa, Van~den Hout, Van~der
  Kooi, Otten, Van~de Vis, et~al.}{Nerbonne
  et~al.}{1996}]{nerbonne1996phonetic}
Nerbonne, J., W.~Heeringa, E.~Van~den Hout, P.~Van~der Kooi, S.~Otten,
  W.~Van~de Vis, et~al. (1996).
\newblock Phonetic distance between dutch dialects.
\newblock In {\em CLIN VI: proceedings of the sixth CLIN meeting}, pp.\
  185--202.

\bibitem[\protect\citeauthoryear{Savage}{Savage}{1957}]{savage1957nonparametric}
Savage, I.~R. (1957).
\newblock Nonparametric statistics.
\newblock {\em Journal of the American Statistical Association\/}~{\em
  52\/}(279), 331--344.

\bibitem[\protect\citeauthoryear{Serva}{Serva}{2021}]{serva}
Serva, M. (2021).
\newblock {Romance language word lists}.
\newblock
  \url{http://people.disim.univaq.it/~serva/languages/55+2.romance.htm}.
\newblock Accessed: 2021-12-23.

\bibitem[\protect\citeauthoryear{Siegel}{Siegel}{1957}]{siegel1957nonparametric}
Siegel, S. (1957).
\newblock Nonparametric statistics.
\newblock {\em The American Statistician\/}~{\em 11\/}(3), 13--19.

\bibitem[\protect\citeauthoryear{Swadesh}{Swadesh}{1955}]{swadesh1955towards}
Swadesh, M. (1955).
\newblock Towards greater accuracy in lexicostatistic dating.
\newblock {\em International journal of American linguistics\/}~{\em 21\/}(2),
  121--137.

\bibitem[\protect\citeauthoryear{Vardi and Zhang}{Vardi and
  Zhang}{2000}]{vardi2000multivariate}
Vardi, Y. and C.-H. Zhang (2000).
\newblock The multivariate l1-median and associated data depth.
\newblock {\em Proceedings of the National Academy of Sciences\/}~{\em
  97\/}(4), 1423--1426.

\bibitem[\protect\citeauthoryear{Wasserman}{Wasserman}{2006}]{wasserman2006all}
Wasserman, L. (2006).
\newblock {\em All of nonparametric statistics}.
\newblock Springer Science \& Business Media.

\bibitem[\protect\citeauthoryear{Wichmann, Rama, and Holman}{Wichmann
  et~al.}{2011}]{wichmann2011phonological}
Wichmann, S., T.~Rama, and E.~W. Holman (2011).
\newblock Phonological diversity, word length, and population sizes across
  languages: The asjp evidence.

\bibitem[\protect\citeauthoryear{Zuo and Serfling}{Zuo and
  Serfling}{2000}]{zuo2000general}
Zuo, Y. and R.~Serfling (2000).
\newblock General notions of statistical depth function.
\newblock {\em Annals of statistics\/}, 461--482.

\end{thebibliography}
\end{document}